\documentclass[11pt]{article}
\usepackage{coling2020}
\usepackage{times}
\usepackage{url}
\usepackage{latexsym}
\usepackage{graphicx}
\usepackage{algorithm}
\usepackage{algorithmic}
\usepackage{multirow}
\usepackage{wrapfig}
\colingfinalcopy 

\title{Inferential Text Generation with Multiple Knowledge Sources and Meta-Learning}

\author{Daya Guo$^1$\thanks{\ \ Equal contribution. Work is done during Daya and Akari's internship at Microsoft Research Asia.}\ , Akari Asai$^{2*}$, Duyu Tang$^{3}$, Nan Duan$^{3}$, Ming Gong$^4$, Linjun Shou$^4$, \\\textbf{Daxin Jiang}$^4$, \textbf{Jian Yin}$^1$ and \textbf{Ming Zhou}$^{3}$\\
$^1$ The School of Data and Computer Science, Sun Yat-sen University. \\Guangdong Key Laboratory of Big Data Analysis and Processing, Guangzhou, P.R.China\\
$^2$ University of Washington\\
$^3$ Microsoft Research Asia, Beijing, China.\\
$^4$ Microsoft Search Technology Center Asia, Beijing, China.\\
}

\date{}

\begin{document}
\maketitle
\begin{abstract}
  We study the problem of generating inferential texts of events for a variety of commonsense like \textit{if-else} relations.
	Existing approaches typically use limited evidence from training examples and learn for each relation individually.
	In this work,  we use multiple knowledge sources 
	as fuels for the model.
	Existing commonsense knowledge bases like ConceptNet  are dominated by taxonomic knowledge (e.g., \textit{isA} and \textit{relatedTo} \mbox{relations}), having a limited number of inferential knowledge.
	We use not only structured commonsense knowledge bases, but also natural language snippets from search-engine results.
	These sources are incorporated into a generative base model via key-value memory network.
	In addition, we introduce a meta-learning based multi-task learning algorithm.
	For each targeted commonsense relation, we regard the learning of examples from other relations as the meta-training process, and the evaluation on examples from the targeted relation as the meta-test process.
	We \mbox{conduct} experiments on Event2Mind and ATOMIC datasets.
	Results show that both  the integration of multiple knowledge sources and the use of the meta-learning algorithm improve the performance.
\end{abstract}

\section{Introduction}
When a daily event such as ``\textit{Peter makes John's coffee}''  occurs, people have the ability to reason about the causes (e.g. ``\textit{Peter wanted to be helpful}'') and effects (e.g. ``\textit{Peter get thanked}'') of the event. 
Inferential text generation is the task that aims to test a computational system's ability on the reasoning of inferential knowledge.
Given a piece of text like ``\textit{Peter makes John's coffee}'' and one of the pre-defined relations like ``\textit{cause}'', the task aims to generate the desired sequence of words.

We study the problem on two benchmark \mbox{datasets}, Event2Mind \cite{P18-1043} and ATOMIC \cite{sap2019atomic}, both of which call for commonsense inference on events.
Till now, sequence-to-sequence model trained on each relation separately achieves the  promising performance on both datasets \cite{sap2019atomic,du2019modeling}.
We contribute at the knowledge source level and at the training algorithm. 
{Firstly}, given a text and a particular relation as input, 
our approach uses automatically retrieved evidences from external multiple knowledge sources, including ConceptNet and search-engine results, for the generation for an event.
{Secondly}, instead of training each relation separately, we regard examples of other relations as auxiliary tasks to improve the targeted relation. We use model-agnostic meta-learning (MAML) \cite{finn2017model} here, which has achieved promising performances on low-resource image classification and reinforcement learning.
We regard learning from other relations as the meta-training process,
and the evaluation on the targeted relation as the meta-test process. 

Results on both datasets show that the integration of external knowledge sources improves the performance, and using multi-task learning with MAML brings further improvements. 

\section{Task and Base Model}

Given an event phrase $x = \{x_1, x_2, \cdots, x_m\}$  and a commonsense relation $c$ as input, the task is to generate 
a sequence $y = \{y_1, y_2, \cdots, y_n\}$, which is the desired hypothesis for the input event on the given relation $c$.

We evaluate on Event2Mind \cite{P18-1043} and ATOMIC \cite{sap2019atomic}, both of which contain about 25,000  event phrases.
Event2Mind focuses on three relations related to mental states (i.e. intents and reactions of the actors), while ATOMIC has nine inferential dimensions includes mental states (the mental pre- and post- conditions of events), event (events about pre- and post- conditions of events) and persona (a stative relation about how the subject of an event is perceived). 

\subsection{Base Model: Encoder-Decoder}
Our base model is an encoder-decoder approach conditioned on a particular relation.

\paragraph{Encoder}
A bi-directional RNN with gated recurrent unit (GRU) \cite{D14-1179} is used to read a event phrase $x=(x_1,...,x_m)$ conditioned on a inference type $c$. Specially, at $i$-th step, we concatenate the embedding of $i$-th word and the inference type $c$ as the input. We then get the final representation of the source sentence $h_x=([\overrightarrow{h_m};\overleftarrow{h_1}])$, where $\overrightarrow{h_m}$ and $\overleftarrow{h_1}$ are last hidden states of the forward and backward RNN, respectively.  


\paragraph{Decoder}
We use a GRU with an attention mechanism as the decoder.
At each time-step $t$, the context vector $c_{t}$ is computed same as the multiplicative attention \cite{luong-pham-manning:2015:EMNLP}.
Afterwards, the concatenation of the context vector, the embedding of the previously predicted word $y_{t-1}$, the embedding of the inference type $h_c$ and the last hidden state $s_{t-1}$ is fed to the next step. After obtaining hidden states $s_{t}$ by GRU, we predict a word from the target vocabulary by a linear layer followed by a softmax function.
\section{Approach}
In this section, we first describe the extraction and the use of knowledge sources, and then describe the use of model-agnostic meta-learning.

\subsection{Knowledge from ConceptNet}
We use triplets $f_i = (subject, relation, object)$ from ConceptNet as knowledge. 
We follow \newcite{wang2018yuanfudao}, and retrieve triples from ConceptNet that contain any n-gram in the target sentence. 
Yet, ConceptNet is dominated by taxonomic knowledge (e.g., ``\textit{river is related to water.}''), and inferential knowledge (e.g., ``\textit{a gift is used for celebrating a birthday}'') tend to be rare.
For instance, among the entire collections of knowledge triplets in ConceptNet, the most frequent relationship type is ``{\it RelatedTo}'' (37.5\%) followed by ``{\it isA}'' (7.0\%), ``{\it AtLocation}'' (5.2\%), and ``{\it Synonym}'' (4.6\%).
We attempts to use search results to increase the coverage.
%

\subsection{Knowledge from Web Search Snippets}

As suggested by \newcite{emami2018generalized}, texts from search engines provide valuable information for commonsense question answering such as Winograd Schema Challenge.
To extract knowledge from web search, we first prepossess an input even phrase $E$, removing place holders and stop words to keep the essential terms of the event. We then automatically generate search queries by concatenating the prepossessed input event phrases and a pre-defined keyword phrase. We use Google search in this work.
The list of part of predefined key phrases for ATOMIC and Event2Mind could be found in Table \ref{table:search-query}. 
As retrieved results contain many irrelevant words, we only retain nouns, adjectives, and verbs.
\begin{table}[h!]\small

  \begin{minipage}[b]{0.48\textwidth} 
    \centering 
	\begin{center}
		\begin{tabular}{l|p{5cm}} \hline
			relation & key phrases  \\ \hline
			xIntent & motivated by, has subevent, intentions, why \\ 
			\{o, x\}React & causes, has subevent, reactions \\ 
			xAttr & has property, attribute, who \\ 
			xNeed & needs, motivated by, has prerequisite, before  \\ 
			\{o, x\}Want & motivated by, causes desire, intentions  \\
			\{o, x\}Effect & causes, has subevents, effects, influences\\
			\hline
		\end{tabular}
		\caption{\label{table:search-query} Examples of key phrase for knowledge hunting from web-search results.}
	\end{center}
	\end{minipage}
	\ \ \ \ \ 
	\begin{minipage}[b]{0.48\textwidth}
	\centering
	\begin{tabular}{p{2.6cm}|cc} \hline
			& ConceptNet &  Web Search \\ \hline
			\# of knowledge per event & \multirow{1}{*}{201}  &  \multirow{1}{*}{3,780} \\
			\hline
			Hit @ xIntent & 28.32\% &    86.1\%
			\\
			Hit @ xReact & 6.76\%  &   63.7\% \\
			Hit @ oReact & 2.37 \% &   57.7\% \\ \hline
		\end{tabular}
		\caption{\label{table:knowledge_coverage} The coverage of existing knowledge bases and the natural language snippets results on web search.}
	\end{minipage}
\end{table}

We randomly sample 100 examples from the Event2Mind development dataset, and calculate the overlap of tokens in each triplet with any of the knowledge triplets.
As shown in Table \ref{table:knowledge_coverage}, the coverage of ConceptNet is low, while the coverage of the triplet extracted from search snippets results is higher.

\subsection{Key-Value Memory}
We treat subjects with relations (search queries) as keys and objects (search results) as values in ConceptNet (Web Search), and retrieved key-value pairs from external knowledge bases are stored in a memory. Similar to \newcite{miller2016key}, we first use source sentence $h_x$ to calculate the relevance between the event phrase and keys through attention mechanism, and then obtain the knowledge representation $h_k$ by weighting averaging values according to the relevance. $h=([h_x;h_c;h_k])$ is used as initial hidden state of the decoder.


\subsection{Multi-Task Learning with Meta-Learning}

We human beings
are very versatile in that we have the ability to leverage experiences learnt from other tasks to help us complete the task at hand. 
In this work, a natural intuition for multi-task learning is to use examples from other relations to improve the targeted relation. 
This can be directly modeled by model-agnostic meta-learning \cite{finn2017model}, which has a meta-train step to quickly update the parameter with several gradient decent steps, followed by a meta-test step which evaluates the new parameter. The final loss at the meta-test step will be used to measure the goodness of the entire learning process and update the model parameter.
We summarize the learning algorithm in Algorithm 1.
In this work, for each targeted relation, we regard the learning of examples from other relations as the meta-train process (line 4-8), and the evaluation on examples from the targeted relation as the meta-test process (line 8). 
Meanwhile, we remain the original supervised loss function for each relations (line 9).

\begin{algorithm}[h]\small
	\caption{}         
	\label{alg1}                          
	\begin{algorithmic}[1]
		\REQUIRE Training datapoints $\mathcal{D} = \{{\rm x^{(j)}, {\rm y^{(j)}}\}}$
		\REQUIRE $\alpha, \beta$: step size hyper parameters
		\STATE Randomly initialize $\theta_0$
		\WHILE{not done} 
		\STATE Sample batch of data $\mathcal{D}_i$ for each $\mathcal{T}_i$ $\in$ $\mathcal{T}$
		\FORALL{$\mathcal{T}_i$}
		\STATE Evaluate $\Delta_{\theta}\sum_j\mathcal{L}_{\mathcal{T}_j} (f_\theta, \mathcal{D}_j)$ using training samples from \{$\mathcal{D}_j$\}$_{j!=i}$ 
		\STATE Compute parameters with gradient descent: $\theta_i^{'} = \theta - \alpha \Delta_{\theta}\sum_j\mathcal{L}_{\mathcal{T}_j} (f_\theta, \mathcal{D}_j)$
		\ENDFOR
		\STATE Update $\theta \Leftarrow \theta - \beta \Delta_{\theta} \mathcal{L}_{\mathcal{T}_i}(f_{\theta_i^{'}}, D^1_i)$ using one batch of data from  task $\mathcal{T}_i$ 
		\STATE Update $\theta \Leftarrow \theta - (1-\beta) \Delta_{\theta}\mathcal{L}_{\mathcal{T}_i}(f_{\theta}, D^2_i)$ using another batch of data from task $\mathcal{T}_i$ 
		\ENDWHILE
	\end{algorithmic}
\end{algorithm}

\section{Experiment}

Table \ref{table:event2mind_results} and Table \ref{table:atomic_bleu} report results of different approaches on Event2Mind and ATOMIC \mbox{datasets}. We follow \cite{P18-1043} and \cite{sap2019atomic} to use Recall and Bleu-2 at top 10 generated texts as evaluation metrics ({\bf Recall@10} and {\bf BLEU@10}), respectively. Training details are given in appendix.

\begin{table*}[h]\small
	\centering
	\begin{tabular}{l|ccc|ccc}
		\hline
		\multirow{2}{*}{Methods}& \multicolumn{3}{c|}{Dev} & \multicolumn{3}{c}{Test} \\
		& {xIntent} & {xReact} & {oReact} & {xIntent} & {xReact} & {oReact}\\
		\hline
		Single-task &42.05\% &43.97\% &69.91\% &42.29\%&44.55\%&69.94\% \\
		Single-task+ConceptNet &43.24\% &44.02\% &70.02\% &43.43\%&44.94\%&69.87\% \\
		Single-task+Google &42.69\% &44.18\% &70.12\% &42.94\%&44.79\%&70.01\% \\
		Single-task+ConceptNet+Google &43.31\% &44.20\% &70.24\% &\bf{43.52}\%&\bf{45.04}\%&\bf{70.25}\% \\
		\hline
		Multi-task+ConceptNet+Google &43.21\% &43.87\% &70.38\% & 43.28\%&45.03\%&69.97\% \\ 	
		MAML+ConceptNet+Google &43.24\% &44.17\% &70.54\% &\bf{43.50}\%&\bf{45.32}\%&\bf{70.52}\% \\ 	
		\hline
	\end{tabular}
	\caption{Recall@10 on three inference types of the Event2Mind dataset with different approaches.}
	\label{table:event2mind_results} 
\end{table*}

{\bf Single-task} training different sequence-to-sequence models for each inference type separately, {\bf Multi-task} represents multi-task learning way, and {\bf ConceptNet} ({\bf Google}) stands for knowledge resources. We can see that incorporating knowledge resources achieve a gain of 0.6\% recall and 0.3\% bleu on Event2Mind and ATOMIC datasets, respectively. Results also show that applying our {\bf MAML} framework, shown in Algorithm \ref{alg1}, to multi-task learning performs better on the majority of inference types. 

\begin{table*}[h]
	\begin{center}
		{\small
			\begin{tabular}{l|ccccccccc} \hline 
				Methods& xIntent &  xNeed & xAttr  & xEffect & xWant & xReact & oEffect& oWant& oReact\\\hline
				%
				%
				%
				%
				%
				9ENC9DEC \cite{sap2019atomic}& 3.47 &  9.93  & 1.64 & 7.53 &7.66 &3.15 &5.02 & 8.12&3.51\\
				
				Single-task & 6.03& 16.85 &  \bf{4.81} & 9.10 &  11.13 & 4.75 & 4.61 & 7.38& 4.41 \\
				Single-task+ConceptNet & 6.21& 16.62&  4.72 & 9.13& 11.20  & 4.62 &4.24  &7.50 & 4.77 \\
				Single-task+Google & 6.45 & 16.86& 4.77  & 9.21& 11.30  &\bf{4.87}  & 4.13 & \bf{8.36}& \bf{4.81} \\
				Single-task+ConceptNet+Google & \bf{6.46} & \bf{17.06}& 4.77  & \bf{9.44}&\bf{11.35}  &4.68  & \bf{4.64} & 6.54& 4.70 \\
				\hline
				Multi-task+ConceptNet+Google&  \bf{7.24}&16.66&  4.81 & 9.57&11.55 &\bf{4.82} & 5.89 & 8.99&\bf{4.71} \\
				
				MAML+ConceptNet+Google & 6.70 & \bf{17.48} & \bf{4.88}  & \bf{9.98}& \bf{11.80}  & 4.64 & \bf{6.26} & \bf{9.02}& 4.36 \\
				\hline
			\end{tabular}
		}
	\end{center}
	\caption{\label{table:atomic_bleu} BLEU@10 on nine inference types of the ATOMIC test dataset with different approaches.}
\end{table*}






\subsection{Effect of the number of search snippets}
\begin{wrapfigure}{r}{0.5\textwidth}
	\centering
	\includegraphics[width=0.49\textwidth]{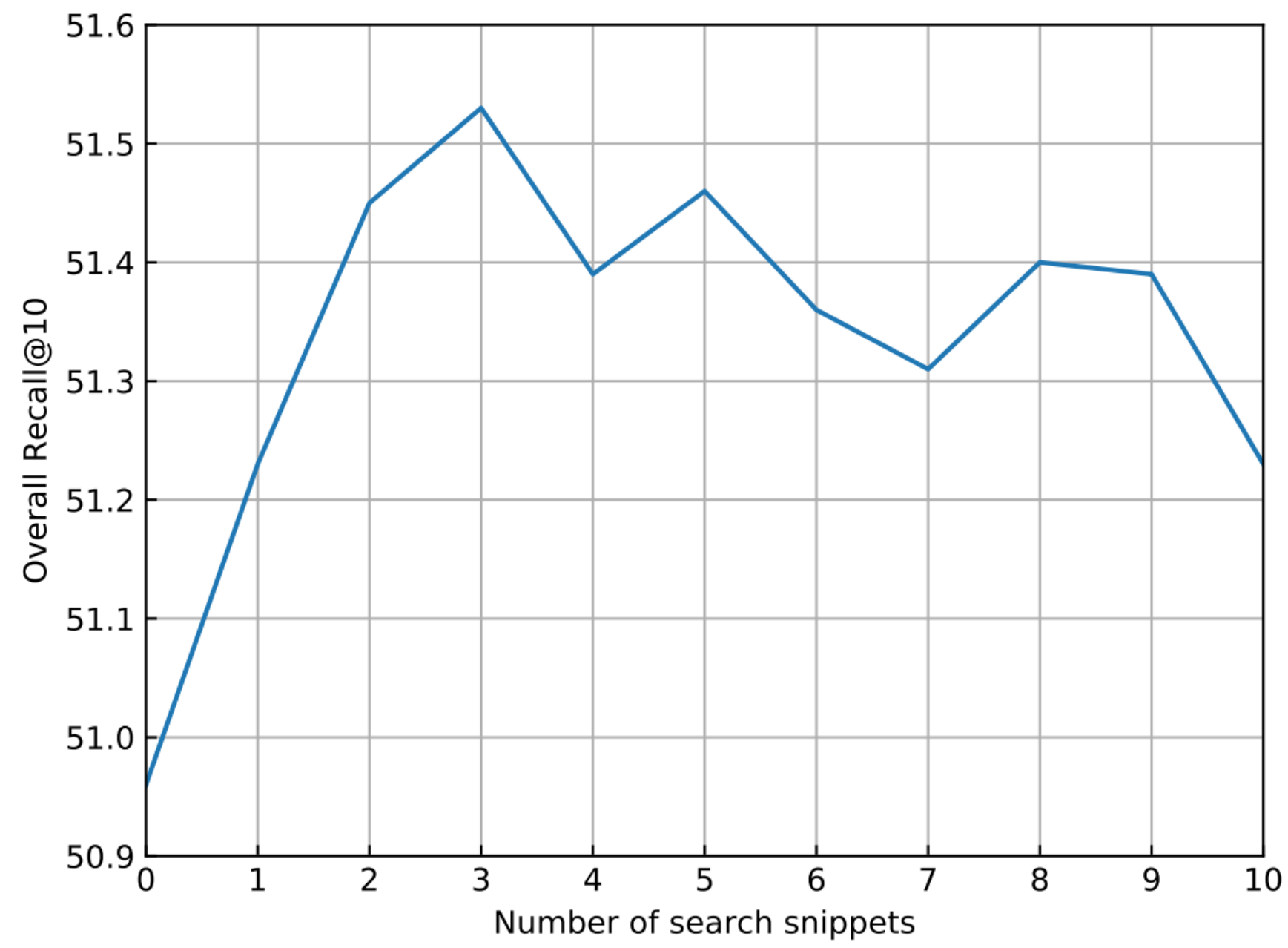}
	\caption{Overall Recall@10 with different number of Google search snippets on the Event2Mind dataset.}
	\label{figure:training_example_num}
\end{wrapfigure}
We study how the number of Google search snippets affects the performance of the model, shown in Figure \ref{figure:training_example_num}. We can see that applying Google search snippets could bring higher recall, which demonstrates the usefulness of knowledge. Although collected search snippets are valuable knowledge sources, they might contain more noise with the increasing number of Google search snippets, which hurts performance of the model.

\subsection{Error analysis}
We analyze randomly selected 100 wrongly predicted instances on the ATOMIC dataset, and summary three main classes of errors. The first problem is that most examples generate correct texts but not in the set of gold answers, which needs more careful evaluation by humans. The second is that the model mistakes inference types. Specially, given a same input and different inference type, the model tends to generate similar outputs. This problem might be mitigated by incorporating more information of inference type. Lastly, some examples fails to generate correct texts since lacking of specific commonsense knowledge. For examples, ``PersonX talks in class'' will be punished but the model generates ``listens to the teacher'' and ``PersonX drinks \_\_\_ everyday" will gain weight but the model generates ``loses weight''. There are two potential directions to make further improvements. The first direction is to leverage more knowledge resources from different dimension. The second direction is to utilize more powerful pre-trained model, such as BERT \cite{devlin2018bert}.

\section{Conclusion}
We present a generative model for generating inferential text of \textit{if-else} relations.
We incorporate two types of knowledge from ConceptNet and Google search results, and use model-agnostic meta-learning (MAML) to utilize examples from other relations.
Experiments show that the integration of external knowledge and MAML both improve the accuracy. We plan to extend the work to the generation of longer text such as essay generation \cite{feng2018topic}. 
\bibliographystyle{coling}
\bibliography{coling2020}

{\bf Appendix A. Dataset}

A brief statistics and comparison of two datasets are given in Table \ref{table:stat_datasets}.
\begin{table}[h]
	\begin{center}
		{
			\begin{tabular}{l|c|c} \hline 
				& Event2Mind & ATOMIC \\ \hline 
				\# of relations & 3 & 9 \\
				\# of events &  24,716 & 24,313 \\
				\# of triplets &  171,291  & 877,108 \\
				\hline
			\end{tabular}
		}
	\end{center}
	\caption{\label{table:stat_datasets} Statistics of Event2Mind~\cite{P18-1043} and ATOMIC~\cite{sap2019atomic}.}
\end{table}

{\bf Appendix B. Model Training}

Here, we list our training details. Word embedding values are initialized with GloVe~\cite{pennington2014glove} and ELMo embeddings~\cite{peters2018deep}. 
We use dropout with a rate of 0.2 for word embeddings and the dimension of the encoder hidden state is 100. 
We set the maximum number of the knowledge triples as 30. 
Model parameters are updated using the Adam method, and the learning rate are 0.0001 and 0.0002 for Event2Mind and ATOMIC datasets, respectively. 
For MAML training, we set the step size $\alpha$ as 0.001, the weight $\beta$ as 0.01 and batch size as 64 for both experiments.  Model hyperparameters are tuned on the development set.
\end{document}


\maketitle

\section{Dataset}

A brief statistics and comparison of two datasets are given in Table \ref{table:stat_datasets}.
\begin{table}[h]
	\begin{center}
		{
			\begin{tabular}{l|c|c} \hline 
				& Event2Mind & ATOMIC \\ \hline 
				\# of relations & 3 & 9 \\
				\# of events &  24,716 & 24,313 \\
				\# of triplets &  171,291  & 877,108 \\
				\hline
			\end{tabular}
		}
	\end{center}
	\caption{\label{table:stat_datasets} Statistics of Event2Mind~\cite{P18-1043} and ATOMIC~\cite{sap2019atomic}.}
\end{table}

\section{Model Training}

Here, we list our training details. Word embedding values are initialized with GloVe~\cite{pennington2014glove} and ELMo embeddings~\cite{peters2018deep}. 
We use dropout with a rate of 0.2 for word embeddings and the dimension of the encoder hidden state is 100. 
We set the maximum number of the knowledge triples as 30. 
Model parameters are updated using the Adam method, and the learning rate are 0.0001 and 0.0002 for Event2Mind and ATOMIC datasets, respectively. 
For MAML training, we set the step size $\alpha$ as 0.001, the weight $\beta$ as 0.01 and batch size as 64 for both experiments.  Model hyperparameters are tuned on the development set.

\bibliographystyle{coling}
\bibliography{coling2020}